%% file: neurips_2026.tex
\documentclass{article}

\PassOptionsToPackage{numbers, compress}{natbib}
\usepackage{pifont}

\newcommand{\cmark}{\ding{51}}

\newcommand{\xmark}{\ding{55}}
\usepackage[preprint]{neurips_2026}

\usepackage[utf8]{inputenc} 
\usepackage[T1]{fontenc}    
\usepackage{hyperref}       
\usepackage{url}            
\usepackage{booktabs}       
\usepackage{amsfonts}       
\usepackage{nicefrac}       
\usepackage{microtype}      
\usepackage{amsmath}
\usepackage{graphicx}
\usepackage{booktabs} 
\usepackage{multirow}

\usepackage[table,dvipsnames]{xcolor}

\title{
SkyVLaM\raisebox{-0.12em}{\hspace{0.2em}\includegraphics[height=0.95em]{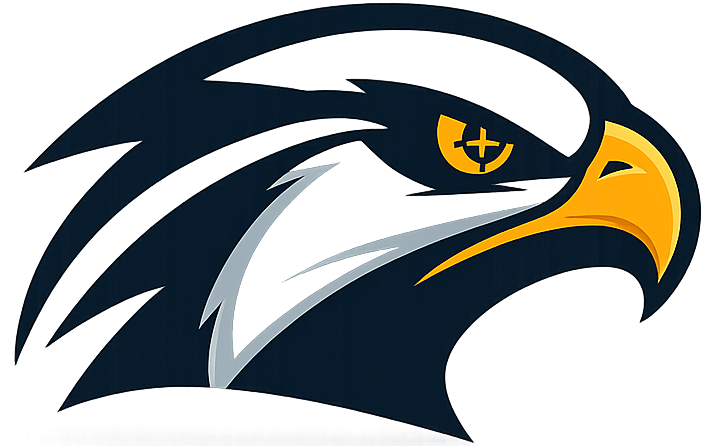}}:
Multimodal Large Language Model for UAV Video Understanding in Remote Sensing
}

\author{%
  Kaiwen Jing$^{1}$\thanks{Equal contribution.} \quad
  Ruixu Jia$^{1}$\footnotemark[1] \quad
  Bingyao Li$^{1}$ \quad
  Ruizhe Ou$^{2}$ \quad
  Ming Wu$^{1}$ \quad
  Chuang Zhang$^{1,3}$\thanks{Corresponding author: \texttt{zhangchuang@bupt.edu.cn}.} \\
  $^{1}$Beijing University of Posts and Telecommunications, Beijing, China\\
  $^{2}$Peking University, Beijing, China\\
  $^{3}$Beijing Wuzi University, Beijing, China
}

\begin{document}

\maketitle

\begin{abstract}
Recent advances in Multimodal Large Language Models (MLLMs) have significantly improved remote sensing (RS) multimodal understanding.
Language-conditioned segmentation is crucial for fine-grained target understanding in Unmanned Aerial Vehicle (UAV) videos. 
However, this task remains challenging due to the prevalence of small, visually ambiguous targets and dynamic aerial perspectives.
In this paper, we propose \textbf{SkyVLaM}, a multimodal large language model for UAV video understanding. 
SkyVLaM constructs sparse tokens directly from patch-level video representations through a temporal basis perceiver, regularizes the sparse basis to encourage complementary temporal cues, and adaptively selects a temporally coherent dense segment for high-resolution inspection. 
The resulting sparse and dense tokens are jointly processed by a large language model for query-conditioned segmentation. 
We further build SkyVid, consisting of SkyVid-VGCG and SkyVid-RVOS for video grounded conversation generation and referring video object segmentation, respectively.
SkyVid contains 101 videos, 33.6K frames, and 1.53M pixel-level object instances.
Experiments show that SkyVLaM provides a more effective allocation of the visual token budget and improves language-conditioned video segmentation in UAV scenarios.

\end{abstract}

\section{Introduction}

\begin{figure}[t]
    \centering
    \includegraphics[width=0.95\textwidth]{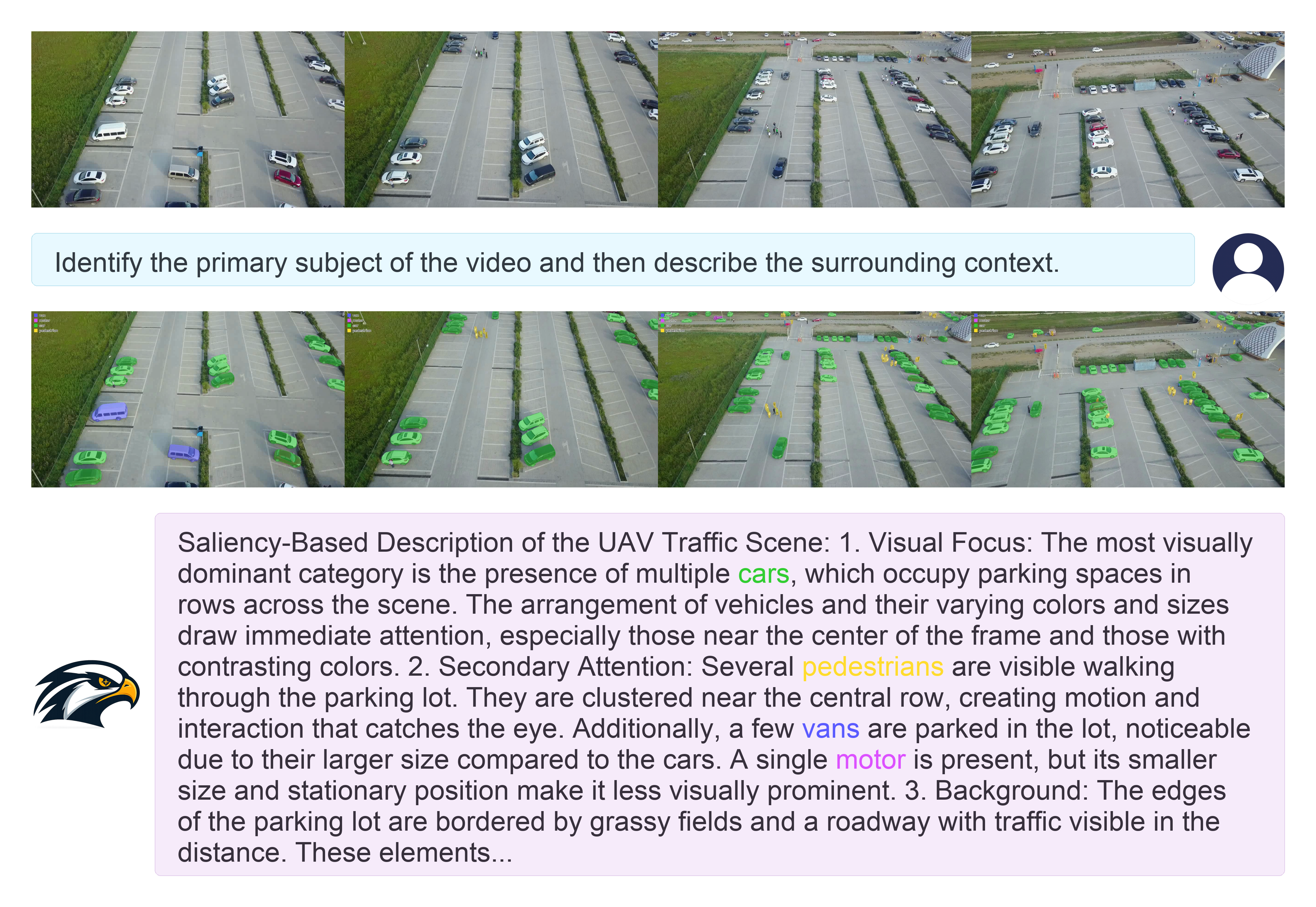}
    \caption{\textbf{Qualitative example of SkyVLaM on UAV video pixel grounding.}
    Given a UAV video and a user instruction, SkyVLaM generates a grounded response while associating referred categories with pixel-level masks.
    The example shows that the model can identify the primary visual subject in the video, describe surrounding contextual objects, and ground multiple categories such as cars, pedestrians, vans, and motors across frames.}
    \label{fig:qualitative_example}
\end{figure}

Multimodal Large Language Models (MLLMs) have demonstrated remarkable performance on remote sensing (RS) multimodal tasks, such as visual question answering (VQA), image captioning, and visual grounding. 
For example, GeoPix \cite{ou2025geopix} adapts a mask predictor to MLLM achieving pixel-level understanding for RS multimodal tasks.
Several studies \cite{irvin2024teochat, liu2024rsunivlm, soni2025earthdial, zhang2025georsmllm} extend remote sensing multimodal models toward multi-temporal reasoning.
However, existing methods primarily focus on static nadir-view satellite imagery or coarse-grained UAV video understanding, and remain limited in instruction-conditioned UAV video analysis with spatio-temporal pixel-level grounding.
Given the rapidly growing demand for UAV-based interaction in urban management, traffic monitoring, and emergency rescue, current RS MLLMs still lack the capability to jointly generate grounded textual responses and precise spatio-temporal masks for referred objects in UAV videos.

UAV applications stand to benefit significantly from pixel-level conversation capabilities. 
While operators typically identify targets via natural language, subsequent analysis often demands precise spatial grounding to move beyond basic pixel-level understanding.
The essence of pixel-level conversation capabilities lies in language-conditioned video segmentation. 
In this setting, a model is tasked with segmenting the referred objects or regions across video frames, guided by natural language instructions.
Unlike coarse-grained detection, classification, or tracking, this task requires aligning language with spatio-temporal visual evidence while maintaining pixel-level precision. 

To this end, we propose SkyVLaM in this study, an RS MLLM for UAV video understanding.
SkyVLaM supports pixel-level video tasks including video grounded conversation generation and referring video object segmentation.
The model integrates precise spatio-temporal masks directly into its conversation, providing grounded textual responses that offer a fine-grained, comprehensive analysis of the video content.
To achieve effective visual token budgeting, we introduce a temporal basis perceiver.
Instead of constructing sparse tokens by independently compressing each frame, SkyVLaM learns a compact sparse representation directly from patch-level video tokens through a temporal basis perceiver. 
The resulting sparse representation contains a global token together with temporally localized basis tokens, allowing the model to preserve both broad scene context and local temporal structure under a constrained token budget. 
To improve the quality of the learned sparse representation, we further introduce a basis diversity regularization that encourages complementary temporal cues rather than redundant summaries.

Based on the learned sparse basis, SkyVLaM then performs adaptive dense selection. 
The key idea is that dense computation should be allocated not by a fixed schedule, but according to what the sparse representation has not yet explained well. 
Concretely, we measure the residual between frame-level visual summaries and the learned basis space, and use it to select a temporally coherent dense segment for high-resolution inspection. 
In this way, sparse tokens provide global temporal abstraction, while dense tokens preserve segmentation-critical spatial details for the most informative part of the clip. 
The resulting sparse and dense tokens are then jointly processed by a large language model for query-conditioned segmentation.

In addition, the remote sensing field lacks object-level textual annotations and pixel-level masks for UAV videos.
To address this gap, we further construct SkyVid, consisting of the SkyVid-VGCG and SkyVid-RVOS subsets, based on VisDrone\cite{zhu2021detection}.
Compared with existing multimodal RS datasets, our datasets more directly reflect the distinctive challenges of UAV videos, encompassing 916 question answering pairs across 101 video sequences.
The videos contain 33.6k frames and 1.53 million objects, with pixel-level mask annotations for every object.
To improve annotation structure and consistency under dynamic UAV perspectives, we design two annotation strategies: Hierarchical Pyramid \& Spatio-temporal Evolution (HPA-STE) and Biomimetic Visual Attention Flow (BVAF).
GPT-4o is used to generate initial textual annotations from the videos and masks, followed by manual review and refinement for scene descriptions, object relations, temporal consistency, and ambiguous expressions.

Experimental results show that SkyVLaM provides an effective token-efficient solution for grounded multimodal understanding in the UAV domain and achieves the best overall performance among the compared methods on the proposed benchmarks.
Our contributions are as follows:
\begin{itemize}
    \item We propose \textbf{SkyVLaM}, an RS MLLM for UAV video understanding, which constructs sparse tokens directly from patch-level video representations, introduces basis diversity regularization to encourage complementary temporal cues, and performs adaptive dense selection to allocate high-resolution computation to the most informative video segment.
    \item We identify sparse-dense token allocation as a central bottleneck in language-conditioned UAV video segmentation, and show that the core challenge lies in balancing broad temporal coverage with segmentation-critical spatial fidelity under a limited visual token budget.
    \item We build the SkyVid-VGCG and SkyVid-RVOS datasets for video grounded conversation generation and referring video object segmentation, with pixel-level masks and textual annotations tailored to  UAV perspectives, and demonstrate that the proposed model provides a more effective and token-efficient solution on both tasks.

\end{itemize}

\section{Related Work}

\subsection{Multimodal Large Language Models}
MLLMs have rapidly evolved from image-level perception and captioning toward temporally aware multimodal reasoning. Representative image-level MLLMs demonstrate strong visual instruction-following and open-ended visual understanding~\cite{li2023blip, liu2023visual}, while video MLLMs further extend these capabilities to multi-frame perception, temporal reasoning, and long-context video understanding~\cite{zhang2023video, maaz2024video, lin2024video}. This shift introduces a new challenge for video modeling: models must preserve temporally relevant visual evidence under limited context and computation.

This development is closely related to referring video object segmentation (RVOS) and grounded video segmentation, which aim to segment or localize target objects in videos according to textual expressions or user instructions. Existing RVOS methods mainly improve cross-modal alignment, temporal propagation, and object-level spatio-temporal reasoning within end-to-end video segmentation architectures~\cite{botach2022end, wu2022language, yan2024referred, ding2023mevis, liang2025referdino}. Motion-centric benchmarks further show that target identity can depend more on temporal dynamics than on static appearance, highlighting the importance of text-driven temporal reasoning for video grounding~\cite{ding2023mevis}. More recently, grounded segmentation and video MLLM methods have extended this paradigm toward image-level reasoning segmentation~\cite{lai2024lisa, ren2024pixellm}, grounded conversation and pixel-level grounding~\cite{rasheed2024glamm, munasinghe2025videoglamm}, language-instructed and reasoning video segmentation~\cite{bai2024one, yan2024visa, gong2025devil, lin2025glus}, and unified image-video dense grounding~\cite{yuan2025sa2va, wei2025instructseg}. While these studies provide important foundations for text-driven dense prediction, they mostly focus on generic visual domains. UAV videos pose a more challenging token allocation problem, since targets are often small, transient, and spatially sparse, and the details required for segmentation may appear only in localized regions.

As video MLLMs scale to longer inputs, visual token allocation becomes a central bottleneck. Existing methods mainly address this issue through query-aware frame selection or visual token compression. Query-aware methods select temporally informative frames using language-aware localizers and keyframe search~\cite{yu2023self, yu2024frame}, adaptive or text-guided sampling~\cite{tang2025adaptive, zhu2025focus, zheng2026tifre}, and reasoning-driven evidence tracing~\cite{zhou2026reason, ghazanfari2025chain}. Token compression methods reduce redundancy through sparse memory and hierarchical representations~\cite{song2024moviechat, weng2024longvlm}, token merging or pruning~\cite{fu2025framefusion, huang2025prunevid}, and structured spatio-temporal compression for long or streaming videos~\cite{hyun2025multi, chen2025streamingtom}. These works show that both temporal coverage and spatial detail preservation are important. For UAV videos, this issue is more critical because small and transient targets require localized segmentation evidence. Unlike generic frame pruning or token compression, our method formulates sparse-dense token budgeting as a learned patch-level temporal basis problem.

\subsection{Remote Sensing Multimodal Large Language Models}
Remote sensing MLLMs have recently progressed from image-text alignment and visual question answering toward instruction following, geospatial dialogue, and pixel-level grounding. 
Early remote sensing MLLMs and benchmarks establish domain-specific image-text understanding~\cite{hu2025rsgpt}, while recent studies improve open-ended dialogue, fine-grained scene reasoning, and remote sensing instruction following~\cite{luo2024skysensegpt, pang2025vhm}. Other works further extend RS MLLMs toward multi-granularity, multi-temporal, and multi-sensory earth observation reasoning~\cite{liu2024rsunivlm, irvin2024teochat, soni2025earthdial, zhang2025georsmllm}. Recent studies also move toward fine-grained RS grounding from complementary directions: GeoPix introduce pixel-level grounding and mask-aware dialogue for remote sensing imagery~\cite{ou2025geopix}, while GeoLLaVA-8K addresses ultra-high-resolution RS understanding through token-efficient modeling~\cite{wang2025geollava}.
SkyEyeGPT further investigates UAV video captioning and visual grounding~\cite{zhan2025skyeyegpt}.
However, existing RS MLLMs remain limited in instruction-conditioned UAV video understanding that jointly requires temporal reasoning, grounded response generation, and spatio-temporal pixel-level mask prediction.
SkyVLaM addresses this setting by supporting both video grounded conversation generation and referring video object segmentation for UAV videos.

\subsection{UAV Benchmarks}
UAV visual benchmarks have also been widely studied, including drone-based detection, tracking, vehicle detection, and cross-modal perception datasets such as VisDrone, UAVDT, and DroneVehicle~\cite{zhu2021detection, du2018unmanned, sun2022drone}. These datasets provide important foundations for UAV object detection, single-object tracking, multi-object tracking, but they generally lack the language annotations, grounded conversations, and pixel-level referring masks required for grounded video understanding. Our work addresses this gap by constructing a low-altitude UAV video dataset tailored to video grounded conversation generation and referring video object segmentation, and by studying language grounding with pixel-level segmentation in dynamic UAV scenes.

\begin{figure}[t]
    \centering
    \includegraphics[width=\textwidth]{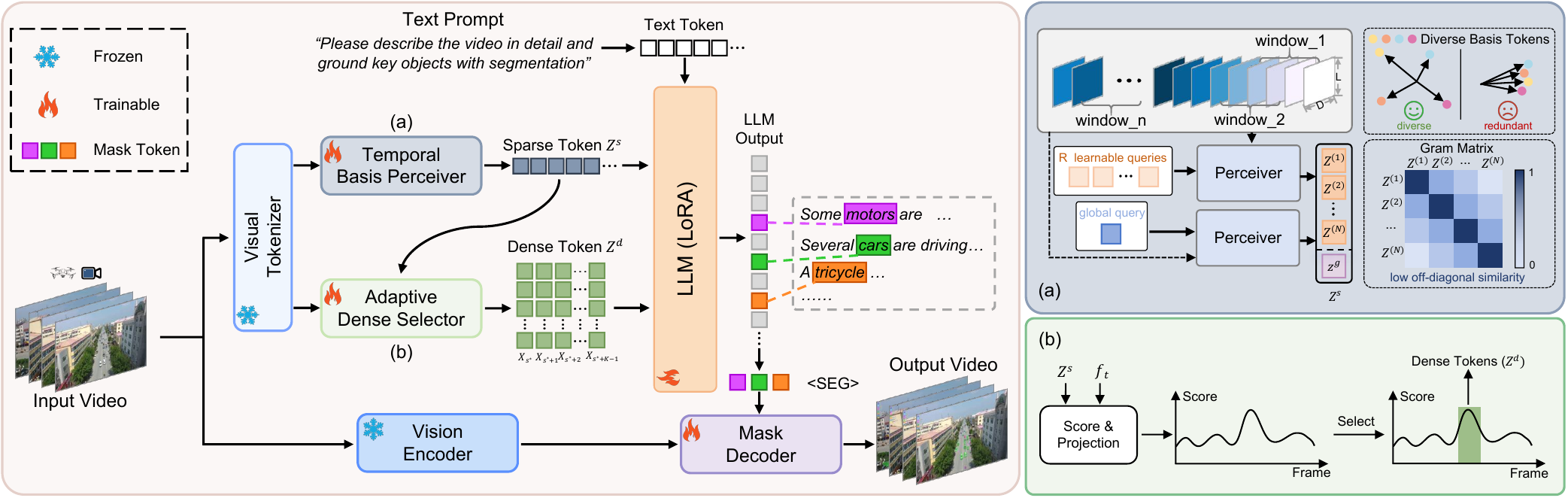}
    \caption{
    Overview of SkyVLaM.
    Given a UAV video and a text prompt, SkyVLaM constructs sparse temporal tokens $\mathbf{Z}^{s}$ and selected dense tokens $\mathbf{Z}^{d}$ for language-conditioned pixel grounding.
    The fused tokens are fed into an LLM to generate grounded responses, and segmentation-related hidden states are decoded into pixel-level masks.
    (a) The temporal basis perceiver summarizes full-video features into global and window-level basis tokens with diversity regularization.
    (b) The adaptive dense selector scores frames against the sparse basis and preserves a continuous informative segment for dense inspection.
    }
    \label{fig:method_overview}
\end{figure}

\section{Method}

\subsection{Overview}
\label{sec:method_overview}

We propose \textbf{SkyVLaM}, a multimodal large language model for UAV video understanding.
Given a UAV video and a language query, SkyVLaM generates a textual response and predicts pixel-level masks for the referred objects or regions.
The model follows a sparse-dense representation design~\cite{bai2024one,xu2024slowfast}: it summarizes the full video with compact temporal basis tokens, preserves dense patch tokens from an adaptively selected segment, and performs language-conditioned mask prediction with the combined visual tokens.
Figure~\ref{fig:qualitative_example} illustrates the task setting and a representative UAV-scene output, where SkyVLaM identifies the primary subject, describes surrounding traffic context, and grounds multiple object categories with pixel-level masks across frames.
The overall architecture of SkyVLaM is shown in Figure~\ref{fig:method_overview}. 
SkyVLaM is built on a video multimodal segmentation pipeline. 
The input video is first encoded into visual tokens, then converted into sparse temporal tokens and selected dense tokens before being fed into the large language model together with the language query.
Specifically, the \emph{temporal basis perceiver} summarizes full-video patch tokens into a compact sparse representation with one global token and multiple local temporal basis tokens, 
while the \emph{adaptive dense selector} uses this sparse basis to select a continuous segment and retain its high-resolution patch tokens. 
The sparse and dense tokens are concatenated as the final video representation, and the large language model produces a grounded response. For mask prediction, segmentation-related hidden states are projected as prompts to a mask decoder and decoded into pixel-level masks. 
The model is initialized from LLaVA~\cite{liu2023visual,rasheed2024llava++} and SAM2~\cite{ravi2024sam}, as detailed in Section~\ref{sec:experimental_setting}.
\subsection{Temporal Basis Perceiver}
\label{sec:temporal_basis}

As illustrated in Figure~\ref{fig:method_overview}(a), we construct sparse video tokens from patch-level video representations using a temporal basis perceiver.
Let $\mathbf{X} \in \mathbb{R}^{T \times L \times D}$ denote the patch-level tokens of a video, where $T$ is the number of frames, $L$ is the number of tokens per frame, and $D$ is the feature dimension.
The temporal basis perceiver produces a compact sparse representation with one global token and multiple local temporal basis tokens.

The global token is generated by applying a learnable global query to the full video token sequence.
It captures clip-level visual context across all frames.
For local temporal basis construction, we divide the video into $N$ overlapping temporal windows.
For the $n$-th window, a set of $R$ learnable latent queries attends to the corresponding patch tokens and produces the window-level basis tokens $\mathbf{Z}^{(n)} \in \mathbb{R}^{R \times D}$.

The final sparse representation is formed by concatenating the global token and all local temporal basis tokens:
\begin{equation}
    \mathbf{Z}^{s} = [\mathbf{z}^{g}; \mathbf{Z}^{(1)}; \mathbf{Z}^{(2)}; \cdots; \mathbf{Z}^{(N)}],
\end{equation}
where $\mathbf{z}^{g}$ is the global token and $\mathbf{Z}^{(n)}$ denotes the basis tokens of the $n$-th temporal window.

To encourage different basis tokens to encode complementary temporal cues, we apply a diversity regularization, as visualized by the Gram-matrix constraint in Figure~\ref{fig:method_overview}(a).
For each temporal window, we normalize its basis tokens and penalize the off-diagonal similarity of their Gram matrix:
\begin{equation}
    \mathcal{L}_{\mathrm{div}} =
    \frac{1}{R^2}
    \left\|
    \widehat{\mathbf{Z}}^{(n)}
    \big(\widehat{\mathbf{Z}}^{(n)}\big)^{\top}
    - \mathbf{I}
    \right\|_{F}^{2},
\end{equation}
where $\widehat{\mathbf{Z}}^{(n)}$ denotes the row-normalized basis token matrix.
The loss is averaged across temporal windows and training samples.
This regularization encourages the sparse tokens to cover diverse temporal evidence under a limited token budget.

\subsection{Adaptive Dense Selection}
\label{sec:adaptive_query}

Sparse basis tokens provide broad temporal coverage, but pixel-level grounding still benefits from high-resolution patch tokens in informative frames.
As shown in Figure~\ref{fig:method_overview}(b), the adaptive dense selector therefore scores each frame by measuring the residual between its frame summary and the sparse basis space, and selects a continuous high-scoring segment for dense inspection.
Given the patch-level video tokens $\mathbf{X}$ and the sparse representation $\mathbf{Z}^{s}$, we first compute a compact frame summary $\mathbf{f}_{t} \in \mathbb{R}^{D}$ for each frame.
We then build a basis bank from the sparse tokens and estimate the residual between each frame summary and its projection onto the sparse basis space:
\begin{equation}
    r_t =
    \left\|
    \mathbf{f}_{t}
    -
    \Pi_{\mathbf{Z}^{s}}(\mathbf{f}_{t})
    \right\|_2^2,
\end{equation}
where $\Pi_{\mathbf{Z}^{s}}(\cdot)$ denotes the projection onto the subspace spanned by the sparse basis tokens.
The residual score $r_t$ measures the complementary visual evidence of frame $t$ with respect to the sparse representation.

We select a continuous segment of $K$ frames according to the accumulated residual score.
For each candidate starting index $s$, the segment score is computed as the sum of residuals within the segment.
The selected segment is given by
\begin{equation}
    s^{*} = \arg\max_{s} \sum_{t=s}^{s+K-1} r_t,
    \qquad
    \mathcal{T}_{d} = \{s^{*}, s^{*}+1, \dots, s^{*}+K-1\}.
\end{equation}
The full patch tokens from the selected frames are retained as dense tokens $\mathbf{Z}^{d}$.

The final video representation is obtained by concatenating the sparse temporal tokens and the selected dense patch tokens:
\begin{equation}
    \widetilde{\mathbf{X}} = [\mathbf{Z}^{s}; \mathbf{Z}^{d}].
\end{equation}
This representation is then used as the visual input to the large language model.

\subsection{Training Objective}
\label{sec:training_objective}

SkyVLaM is trained with language supervision, mask supervision, and basis diversity regularization.
The language modeling loss supervises the generated textual responses.
The segmentation loss supervises the predicted masks using binary cross-entropy and Dice loss:
\begin{equation}
    \mathcal{L}_{\mathrm{seg}} =
    \lambda_{\mathrm{bce}} \mathcal{L}_{\mathrm{bce}} +
    \lambda_{\mathrm{dice}} \mathcal{L}_{\mathrm{dice}}.
\end{equation}
The final objective is
\begin{equation}
    \mathcal{L} =
    \lambda_{\mathrm{lm}} \mathcal{L}_{\mathrm{lm}} +
    \mathcal{L}_{\mathrm{seg}} +
    \lambda_{\mathrm{div}} \mathcal{L}_{\mathrm{div}}.
\end{equation}
Here, $\mathcal{L}_{\mathrm{lm}}$ denotes the language modeling loss, $\mathcal{L}_{\mathrm{seg}}$ denotes the mask prediction loss, and $\mathcal{L}_{\mathrm{div}}$ denotes the basis diversity regularization.

\section{Dataset}

\begin{figure}[t]
    \centering
    \includegraphics[width=\textwidth]{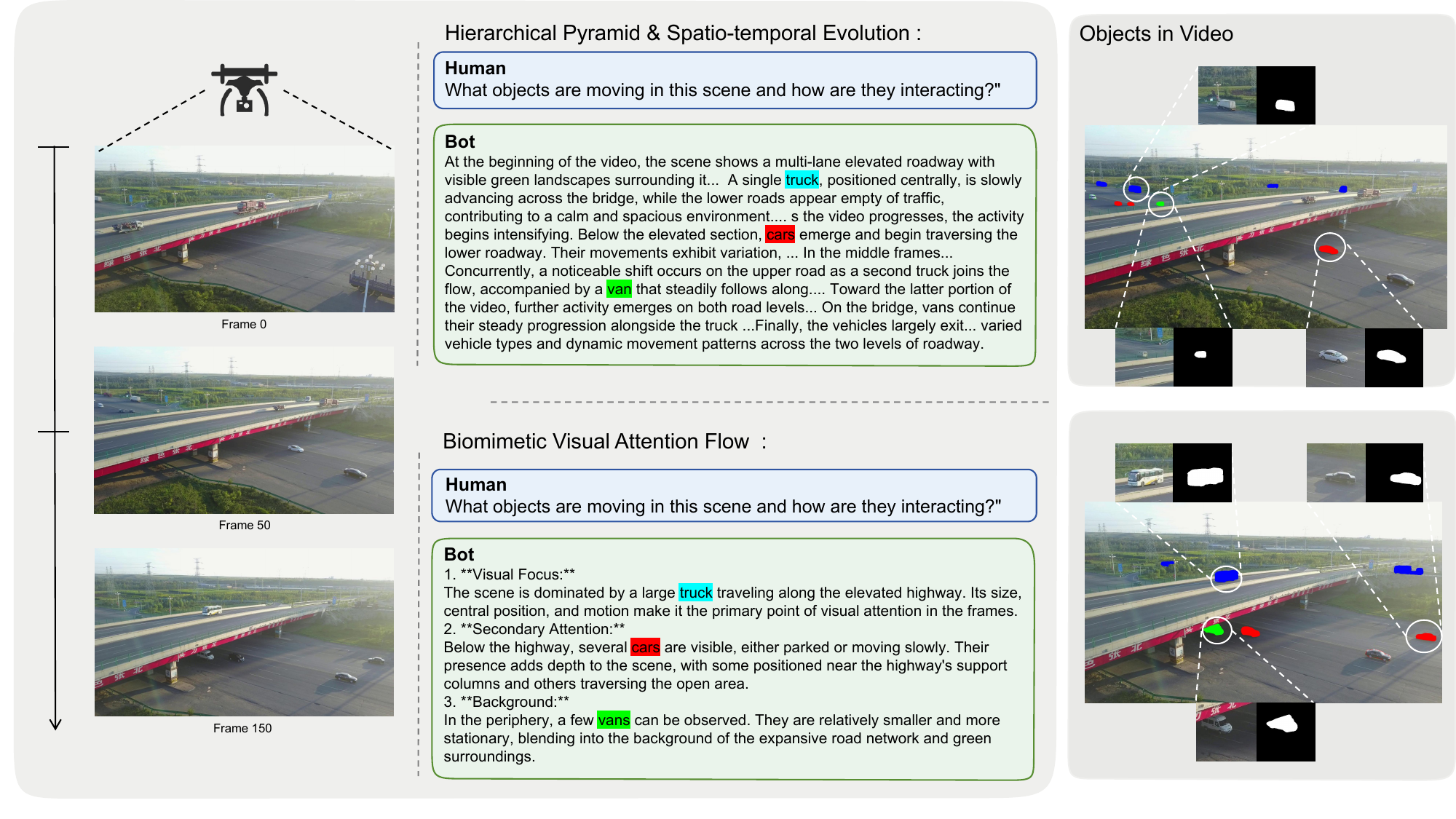}
    \caption{\textbf{Examples of SkyVid-VGCG.}
    The left panel displays representative frames sampled from a UAV video sequence.
    The upper and lower parts of the center panel present textual descriptions generated under the Hierarchical Pyramid \& Spatio-temporal Evolution (HPA-STE) and Biomimetic Visual Attention Flow (BVAF) strategies, respectively.
    The right panel visualizes representative object masks, with blue, red, and green indicating trucks, cars, and vans, respectively.
    }
    \label{fig:dataset_examples}
\end{figure}

In this section, we introduce SkyVid-VGCG (Video Grounded Conversation Generation) and SkyVid-RVOS (Referring Video Object Segmentation), which are specifically designed to support spatio-temporal pixel-level understanding in UAV videos.
The proposed datasets are built upon VisDrone\cite{zhu2021detection}, a large scale benchmark designed for UAV domain. 
We supplemented the VisDrone dataset with pixel-level textual
descriptions.
Our data annotation process consists of two primary stages.
We first augment the VisDrone bounding-box annotations with SAM2-generated object masks to construct SkyVid-RVOS.
Building upon these masks, we then construct SkyVid-VGCG with grounded textual annotations.
The detailed construction pipeline is presented below.

\subsection{SkyVid-RVOS}
\label{subsec:SkyVid-RVOS}

Starting from the VisDrone Multi-Object Tracking benchmark~\cite{zhu2021detection}, we construct SkyVid-RVOS as the mask-level grounding subset of SkyVid.
We feed each video and its associated bounding-box annotations into SAM2~\cite{ravi2024sam} to generate object masks.
To ensure annotation quality, expert annotators manually review the generated masks and correct inaccurate object boundaries, missing or fragmented objects, identity drift, and temporally inconsistent masks across frames.

We then formulate category-oriented referring queries, such as ``Can you segment the truck in this video?'', and adopt a standardized response template, ``It is \textless SEG\textgreater.'', for referring video object segmentation.

\subsection{SkyVid-VGCG}
\label{subsec:SkyVid-VGCG}

Building upon the object masks constructed in SkyVid-RVOS, we further construct SkyVid-VGCG for video grounded conversation generation.
SkyVid-VGCG provides scene- and motion-aware video captions, object-level descriptions, and inter-object relationship descriptions grounded in the corresponding spatio-temporal masks.
To improve the structure and consistency of textual annotations, we design two annotation strategies:

\textbf{Hierarchical Pyramid \& Spatio-temporal Evolution (HPA-STE).}
This strategy organizes descriptions hierarchically from global scene contexts to local object behaviors.
The annotations first describe the overall scene and then progressively focus on object interactions and local activities.
To characterize temporal evolution, indicators such as ``beginning'', ``middle'', and ``end'' are incorporated to describe object appearance, interaction, motion changes, and departure events throughout the video sequence.

\textbf{Biomimetic Visual Attention Flow (BVAF).}
This strategy organizes object descriptions according to a visual-attention order inspired by human visual observation.
It considers factors such as object size, motion intensity, and spatial proximity to the image center.
Objects with higher visual saliency are described first as the visual focus, followed by secondary objects and surrounding background context.
This ordering improves annotation consistency in dense UAV scenes containing small or weakly salient objects.

Based on these strategies, we construct prompts for GPT-4o to generate initial textual annotations from the input videos and masks.
The generated annotations are then manually reviewed to verify scene descriptions, object relations, and temporal consistency, and inaccurate or ambiguous expressions are further refined.

Representative examples are shown in Fig.~\ref{fig:dataset_examples}.
The HPA-STE annotations describe the scene hierarchically from global context to local object behaviors and use temporal indicators to capture object evolution throughout the video.
The BVAF annotations organize objects into visual focus, secondary attention, and background context according to their visual saliency.

\begin{figure}[t]
    \centering
    \includegraphics[width=\textwidth]{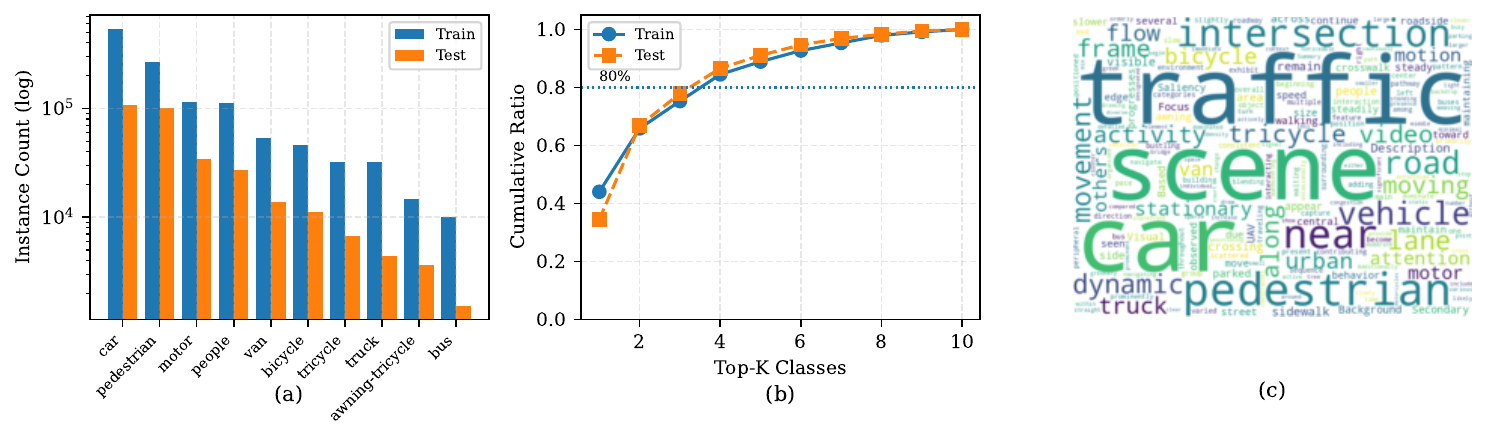}
    \caption{\textbf{Video objects statistics and word cloud.} 
    (a) Objects count distribution across all categories.
    (b) Cumulative distribution function of instance counts.
    (c) Word cloud of textual annotations.
    }
    \label{fig:dataset_dist}
\end{figure}

\subsection{Statistics}
In total, SkyVid contains 101 UAV video sequences, 33.6K frames, and 1.53M object instances annotated with pixel-level masks, yielding an average density of 45.6 instances per frame.
SkyVid-RVOS contains 714 question-answering pairs, while SkyVid-VGCG contains 202 question-answering pairs, resulting in 916 pairs in total.
For dataset splitting, we follow the original VisDrone-MOT partition by merging its training and validation splits into the SkyVid training set and retaining its test split as the SkyVid test set.
After preprocessing, the training set contains 81 video sequences and the test set contains 20 video sequences.
QA pairs are partitioned according to the corresponding video-level split.
We conduct a statistical analysis of the dataset, with the results presented in Fig.~\ref{fig:dataset_dist}.
Specifically, Fig.~\ref{fig:dataset_dist}(a) shows that objects count distribution across all categories.
It can be observed that a small number of head categories (e.g., car and pedestrian) dominate the dataset, while numerous tail classes have significantly fewer samples.
This observation is further supported by the cumulative distribution in Fig.~\ref{fig:dataset_dist}(b), where the top 4 categories already contribute more than 80\% of the total instances.
Such a skewed distribution poses challenges for learning balanced representations and may lead to biased predictions toward frequent classes.
Furthermore, the word cloud in Fig.~\ref{fig:dataset_dist}(c) provides complementary insights into the semantic distribution of the dataset.

\section{Experiments}
\subsection{Experimental Setting}
\label{sec:experimental_setting}
In our implementation, the visual tokenizer and LLM are initialized from LLaVA~\cite{liu2023visual,rasheed2024llava++}, while the vision encoder and mask decoder are initialized from SAM2~\cite{ravi2024sam}.
The trainable components include the LoRA parameters of the LLM, the mask decoder, the visual and language projectors, the temporal basis perceiver, and the adaptive dense selector.
We train SkyVLaM on the proposed SkyVid dataset with a learning rate of $3\times10^{-4}$ and a batch size of 8 for 1,250 optimization steps.
The training data are repeated to form 5,000 samples per epoch.
All experiments are conducted on 8 NVIDIA A100 GPUs.

\begin{figure}[t]
    \centering
    \includegraphics[width=\textwidth]{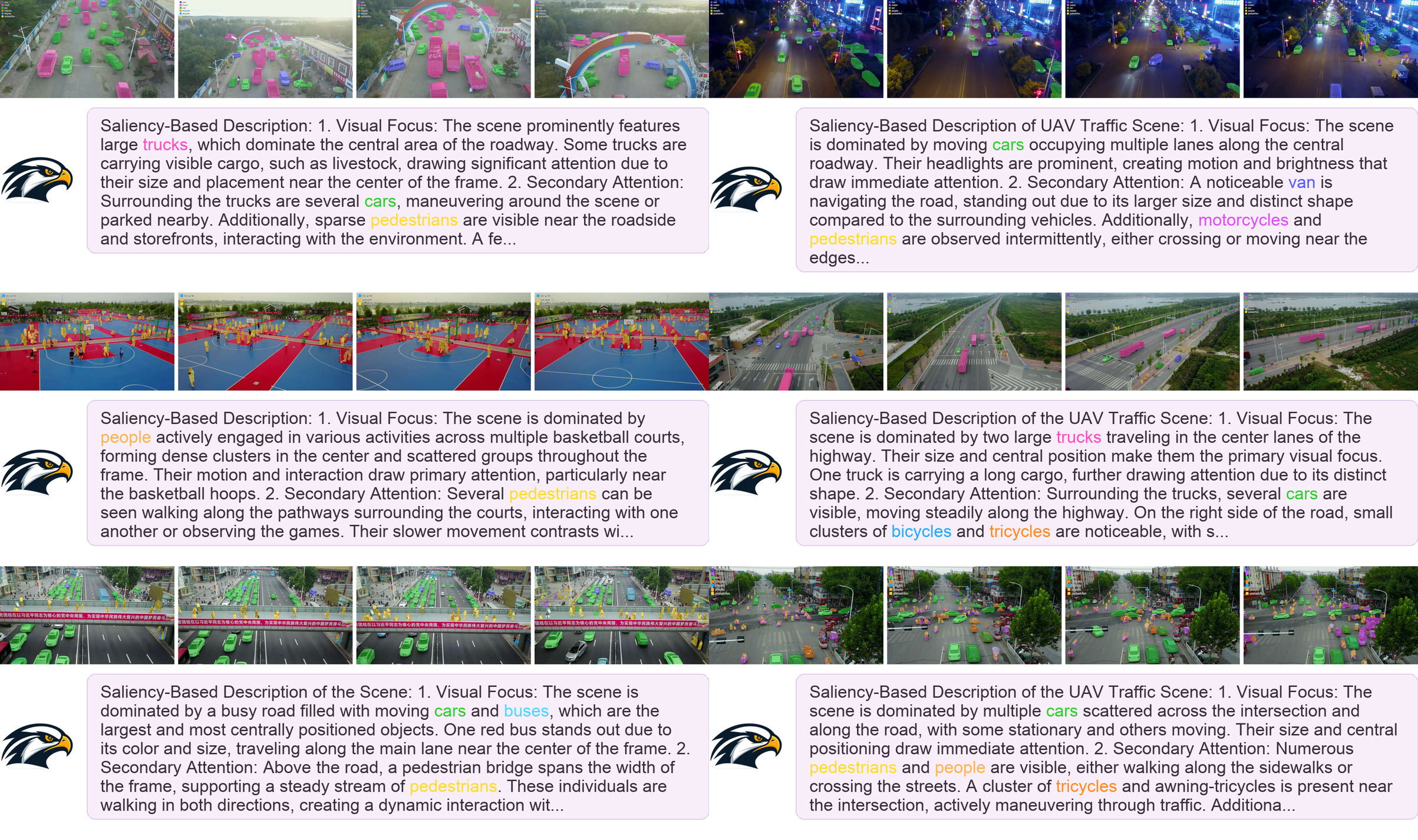}
    \caption{Qualitative results of SkyVLaM on SkyVid-VGCG.}
    \label{fig:qualitative_gcg}
\end{figure}
\input{tables/main_gcg}

\subsection{Video Grounded Conversation Generation}
The video grounded conversation generation task aims to bridge natural language descriptions with fine-grained visual localization through phrase-level grounding. 
The task anchors specific phrases to unique segmentation masks within the video frames. Operationally, we elicit these grounded responses by querying the model for "interleaved segmentation masks". 
The model then generates sequences where referring expressions are delineated by special tokens <p>, </p>, and <SEG> to signify the start, end, and associated mask of each grounded entity. 
This structured output format facilitates the seamless integration of descriptive narratives and precise object masks.

We compare SkyVLaM with VideoGLaMM~\cite{munasinghe2025videoglamm}, VideoLISA~\cite{bai2024one} on SkyVid-VGCG.
To evaluate the quality of the generated content, we employ METEOR~\cite{banerjee2005meteor} and CIDEr ~\cite{vedantam2015cider} as the primary metrics for conversational generation.
As shown in Table~\ref{tab:main_gcg}, our model excels in generating detailed captions with precise object grounding, as evidenced by superior METEOR~\cite{banerjee2005meteor} scores. 
CIDEr~\cite{vedantam2015cider} score is 82.6, significantly outperforming existing methods.
Regarding mask quality, our model consistently outperforms baselines in gIoU, cIoU and mIoU, demonstrating higher spatial alignment with ground truth.
Qualitative results of SkyVLaM on SkyVid-VGCG are shown in Fig.~\ref{fig:qualitative_gcg}.

\subsection{Referring Video Object Segmentation}
The target of referring video object segmentation is to anchor natural language descriptions to specific visual objects across time. 
To probe the spatio-temporal localization performance in dynamic scenarios, SkyVLaM is prompted with fundamental instructions (e.g., "Can you segment the {Object} in this video?").
This setup benchmarks the model’s capability to interpret and ground diverse linguistic cues.
We compare SkyVLaM with VISA~\cite{yan2024visa}, VideoGLaMM~\cite{munasinghe2025videoglamm},  VRS-HQ~\cite{gong2025devil}, VideoLISA~\cite{bai2024one} on SkyVid-RVOS. As shown in Table~\ref{tab:main_rs}, our proposed model delivers the highest performance on the majority of metrics on SkyVid-RVOS.

\input{tables/main_rs}

\subsection{Ablation Studies}
\label{sec:ablation}

We conduct ablation studies on SkyVid-RVOS to examine the main design choices of SkyVLaM.
The baseline follows the sparse-dense sampling strategy of VideoLISA~\cite{bai2024one}: dense frames are uniformly sampled and preserved with full-resolution features, while the remaining interleaved frames are downsampled as sparse tokens.
For a fair comparison, we replace the original SAM decoder with SAM2 while keeping the same visual tokenizer, language model, and training protocol.

\input{tables/ab_b1}

\paragraph{Main component analysis.}
We first isolate the three main components of SkyVLaM: temporal basis construction, basis diversity regularization, and adaptive dense selection.
As shown in Table~\ref{tab:ablation_main}, replacing the VideoLISA-style sparse representation with the temporal basis perceiver improves all metrics, showing that patch-level temporal basis construction provides a stronger sparse representation for UAV video grounding.
Basis diversity regularization and adaptive dense selection further improve the model in different ways.
The diversity regularizer brings a larger gain on mIoU, suggesting improved average mask quality through more complementary temporal basis tokens.
Adaptive dense selection brings stronger gains on gIoU and cIoU, indicating that residual-guided dense inspection improves global grounding and region coverage.
Combining all components gives the best result, reaching 23.52 gIoU, 40.42 cIoU, and 25.87 mIoU.

\input{tables/ab_b2}

\paragraph{Sparse representation design.}
We next examine how sparse video tokens should be constructed.
In Table~\ref{tab:ab_sparse_design}, fixed-budget variants compress full-video patch tokens into the same number of sparse tokens $N_s$, while our temporal basis constructs $1+NR$ sparse tokens, where $N \approx T/S$.
Frame-level compression gives the weakest performance because spatial patch evidence is removed before sparse token construction.
Patch-level variants perform better, and the proposed temporal basis achieves the best result with 21.74 gIoU, 38.26 cIoU, and 23.86 mIoU.
Its compression factor is approximately $TL/(1+NR) \approx SL/R$, which remains controlled by the window stride and basis capacity rather than increasing directly with video length.

\begin{table}[t]
\centering
\footnotesize
\renewcommand{\arraystretch}{1.08}

\begin{minipage}[t]{0.49\linewidth}
\centering
\caption{Ablation of basis diversity regularization on SkyVid-RVOS. Higher is better for all metrics. Best results are shown in \textbf{bold}.}
\label{tab:ab_div_reg}
\vspace{0.3em}
\resizebox{\linewidth}{!}{%
\input{tables/ab_b3}
}
\end{minipage}
\hfill
\begin{minipage}[t]{0.49\linewidth}
\centering
\caption{Ablation of dense selection strategies on SkyVid-RVOS. Higher is better for all metrics. Best results are shown in \textbf{bold}.}
\label{tab:ab_dense_query}
\vspace{0.3em}
\resizebox{\linewidth}{!}{%
\input{tables/ab_b4}
}%
\end{minipage}

\end{table}

\paragraph{Basis diversity regularization.}
We then study whether the learned basis tokens benefit from explicit diversity constraints.
The results in Table~\ref{tab:ab_div_reg} show that all regularized variants improve over the model without regularization.
Window-wise decorrelation performs best, reaching 22.08 gIoU, 38.71 cIoU, and 24.64 mIoU.
This supports the local regularization design, since redundancy is most likely to occur among basis tokens attending to the same temporal window.

\paragraph{Dense selection strategy.}
We further compare different ways of selecting dense frames under the same dense budget $K$.
As reported in Table~\ref{tab:ab_dense_query}, content-agnostic strategies such as random sampling and middle-segment selection are weaker than uniform sampling.
Residual-based selection improves performance, indicating that the sparse basis provides useful signals for locating frames that require dense inspection.
Selecting a contiguous segment performs better than independent top-$K$ selection, showing the importance of local temporal coherence.
Our adaptive contiguous segment achieves the best result, with 22.83 gIoU, 39.64 cIoU, and 24.48 mIoU.

\input{tables/ab_b5}

\paragraph{Sparse-dense token budget.}
Finally, we analyze the trade-off between sparse basis capacity and dense-frame budget.
Table~\ref{tab:ablation_budget} shows that increasing the sparse capacity from $1+2N$ to $1+4N$ improves performance under the same dense-frame budget, and increasing dense frames from 2 to 6 further improves pixel-level grounding.
The default setting, $1+4N$ sparse tokens with 6 dense frames, achieves the best trade-off.
Further increasing the sparse capacity to $1+8N$ slightly reduces performance, suggesting that excessive sparse tokens may introduce redundant visual evidence.
We therefore use $1+4N$ sparse tokens and 6 dense frames as the default configuration.

\section{Conclusion}
In this paper, we study language-conditioned video segmentation in UAV videos from the perspective of sparse-dense visual token allocation. We propose a temporal basis learning framework that constructs sparse tokens directly from patch-level video representations, regularizes the sparse basis to encourage complementary temporal cues, and adaptively selects a temporally coherent dense segment for high-resolution inspection. This design explicitly separates video-level temporal abstraction from frame-level dense inspection, leading to a more effective use of the visual token budget for query-conditioned segmentation.

To support this setting, we further build SkyVid, consisting of the SkyVid-VGCG and SkyVid-RVOS subsets for video grounded conversation generation and referring video object segmentation. Experimental results on both tasks show that the proposed method improves language-conditioned video segmentation in UAV scenarios and provides a stronger sparse-dense formulation than conventional token compression strategies. We hope this work can provide a useful step toward grounded multimodal video understanding in aerial and remote sensing environments.

\section*{Acknowledgments}
This work was supported by NSFC under Grant U24B20177.

\newpage
\bibliographystyle{plainnat}
\bibliography{references}

\end{document}

%% file: tables/main_gcg.tex
\begin{table}[t]
    \centering
    \footnotesize
    \setlength{\tabcolsep}{5pt}
    \renewcommand{\arraystretch}{1.12}
    \caption{Evaluation results on video grounded conversation generation. Higher is better for all metrics. Best results are shown in \textbf{bold}, and second-best results are underlined.}
    \label{tab:main_gcg}
    \begin{tabular}{lccccc}
        \toprule
        \multirow{2}{*}{\textbf{Method}} 
        & \multicolumn{3}{c}{\textbf{Mask Quality}} 
        & \multicolumn{2}{c}{\textbf{Caption Quality}} \\
        \cmidrule(lr){2-4} \cmidrule(lr){5-6}
        & \textbf{gIoU} & \textbf{cIoU} & \textbf{mIoU} & \textbf{CIDEr} & \textbf{METEOR} \\
        \midrule
        VideoGLaMM~\cite{munasinghe2025videoglamm} 
        & 40.08 & 40.19 & 20.19 & \underline{75.1} & \underline{84.44} \\

        VideoLISA~\cite{bai2024one} + SAM2~\cite{ravi2024sam} 
        & \underline{57.48} & \underline{58.14} & \underline{35.65} & 70.1 & 84.42 \\
        
        \rowcolor{gray!10}
        \textbf{SkyVLaM} 
        & \textbf{58.94} & \textbf{60.96} & \textbf{36.67} & \textbf{82.6} & \textbf{86.73} \\
        \bottomrule
    \end{tabular}
\end{table}

%% file: tables/main_rs.tex
\begin{table}[t]
    \centering
    \footnotesize
    \setlength{\tabcolsep}{5pt}
    \renewcommand{\arraystretch}{1.12}
    \caption{Evaluation results on referring video object segmentation. Higher is better for all metrics. Best results are shown in \textbf{bold}, and second-best results are underlined.}
    \label{tab:main_rs}
    \begin{tabular}{p{4.2cm}ccc}
        \toprule
        \textbf{Method} & \textbf{gIoU} & \textbf{cIoU} & \textbf{mIoU} \\
        \midrule
        VISA~\cite{yan2024visa} 
        & 12.10 & 19.70 & 12.10 \\

        VideoGLaMM~\cite{munasinghe2025videoglamm} 
        & 19.60 & 15.60 & 8.80 \\

        VRS-HQ~\cite{gong2025devil} 
        & 14.25 & \underline{39.24} & 14.72 \\

        VideoLISA~\cite{bai2024one} + SAM2~\cite{ravi2024sam} 
        & \underline{20.68} & 37.31 & \underline{22.91} \\

        \rowcolor{gray!10}
        \textbf{SkyVLaM} 
        & \textbf{23.52} & \textbf{40.42} & \textbf{25.87} \\
        \bottomrule
    \end{tabular}
\end{table}

%% file: tables/ab_b1.tex
\begin{table}[t]
    \centering
    \footnotesize
    \setlength{\tabcolsep}{7pt}
    \renewcommand{\arraystretch}{1.12}
    \caption{Main component ablation on SkyVid-RVOS. ``Temp. Basis'' denotes the temporal basis perceiver, ``Div. Reg.'' denotes basis diversity regularization, and ``Adaptive Dense'' denotes adaptive dense selection. Higher is better for all metrics. Best results are shown in \textbf{bold}.}
    \label{tab:ablation_main}
    \begin{tabular}{ccc|ccc}
        \toprule
        \textbf{Temp. Basis} & \textbf{Div. Reg.} & \textbf{Adaptive Dense}
        & \textbf{gIoU} & \textbf{cIoU} & \textbf{mIoU} \\
        \midrule
        \xmark & \xmark & \xmark & 20.68 & 37.31 & 22.91 \\
        \cmark & \xmark & \xmark & 21.74 & 38.26 & 23.86 \\
        \cmark & \cmark & \xmark & 22.08 & 38.71 & 24.64 \\
        \cmark & \xmark & \cmark & 22.83 & 39.64 & 24.48 \\
        \rowcolor{gray!10}
        \cmark & \cmark & \cmark & \textbf{23.52} & \textbf{40.42} & \textbf{25.87} \\
        \bottomrule
    \end{tabular}
\end{table}

%% file: tables/ab_b2.tex
\begin{table}[t]
    \centering
    \footnotesize
    \setlength{\tabcolsep}{4.2pt}
    \renewcommand{\arraystretch}{1.14}
    \caption{Ablation of sparse representation design on SkyVid-RVOS. 
    ``Comp. factor'' denotes the compression factor from original full patch tokens to sparse tokens.
    Here, $T$ is the number of frames, $L$ is the number of patch tokens per frame, $N_s$ is a fixed sparse-token budget, $N \approx T/S$ is the number of temporal windows, $S$ is the window stride, and $R$ is the number of basis tokens per window.
    Higher is better for all metrics. Best results are shown in \textbf{bold}, and second-best results are underlined.}
    \label{tab:ab_sparse_design}
    \begin{tabular}{lccccc}
        \toprule
        \textbf{Sparse representation} 
        & \textbf{Sparse tokens} 
        & \textbf{Comp. factor}
        & \textbf{gIoU} 
        & \textbf{cIoU} 
        & \textbf{mIoU} \\
        \midrule
        Frame mean pooling 
        & $N_s$ 
        & $TL/N_s$ 
        & 20.68 & 37.31 & 22.91 \\
        
        Frame-level MLP projection 
        & $N_s$ 
        & $TL/N_s$ 
        & 20.94 & 37.86 & 23.18 \\
        
        Uniform temporal window pooling 
        & $N_s$ 
        & $TL/N_s$ 
        & 21.16 & 38.04 & 23.35 \\
        
        Patch-level global tokens 
        & $N_s$ 
        & $TL/N_s$ 
        & 21.31 & 37.92 & 23.41 \\
        
        Patch-level local window tokens only 
        & $NR$ 
        & $\approx SL/R$ 
        & \underline{21.52} & \underline{38.11} & \underline{23.59} \\
        
        \rowcolor{gray!10}
        \textbf{Patch-level temporal basis (ours)} 
        & $\mathbf{1+NR}$ 
        & $\mathbf{\approx SL/R}$ 
        & \textbf{21.74} & \textbf{38.26} & \textbf{23.86} \\
        \bottomrule
    \end{tabular}
\end{table}

%% file: tables/ab_b3.tex
    \begin{tabular}{lcccc}
        \toprule
        \textbf{Regularization variant} & $\lambda_{\mathrm{div}}$ & \textbf{gIoU} & \textbf{cIoU} & \textbf{mIoU} \\
        \midrule
        None & 0 & 21.74 & 38.26 & 23.86 \\
        Global decorrelation & 0.05 & 21.91 & 38.44 & 24.19 \\
        Window-wise decorrelation & 0.01 & 22.01 & 38.58 & 24.43 \\
        Cosine dispersion & 0.05 & 21.86 & 38.37 & 24.25 \\
        Orthogonality penalty & 0.05 & 21.95 & 38.51 & 24.32 \\
        \rowcolor{gray!10}
        \textbf{Window-wise decorrelation (ours)} & \textbf{0.05} & \textbf{22.08} & \textbf{38.71} & \textbf{24.64} \\
        \bottomrule
    \end{tabular}

%% file: tables/ab_b4.tex
\begin{tabular}{lccc}
    \toprule
    \textbf{Dense query strategy} & \textbf{gIoU} & \textbf{cIoU} & \textbf{mIoU} \\
    \midrule
    Uniform sampling                      & 21.74 & 38.26 & 23.86 \\
    Random sampling                       & 21.32 & 37.84 & 23.41 \\
    Middle segment                        & 21.58 & 38.09 & 23.72 \\
    Top-$K$ independent by residual       & 22.36 & 39.21 & 24.21 \\
    \rowcolor{gray!10}
    \textbf{Adaptive contiguous segment (ours)} & \textbf{22.83} & \textbf{39.64} & \textbf{24.48} \\
    \bottomrule
\end{tabular}

%% file: tables/ab_b5.tex
\begin{table}[t]
    \centering
    \footnotesize
    \setlength{\tabcolsep}{7pt}
    \renewcommand{\arraystretch}{1.12}
    \caption{Performance under different sparse-dense token budgets on SkyVid-RVOS. Here, $N=8$ denotes the number of temporal windows, and each dense frame contains 576 visual tokens. ``Total tokens'' denotes the total number of visual tokens fed to the multimodal model. Higher is better for all metrics. Best results are shown in \textbf{bold}.}
    \label{tab:ablation_budget}
    \begin{tabular}{ccccc}
        \toprule
        \textbf{Sparse tokens} & \textbf{Dense frames} & \textbf{Total tokens} & \textbf{gIoU} & \textbf{mIoU} \\
        \midrule
        $1 + 2N$ & $2$ & 1169 & 22.31 & 24.57 \\
        $1 + 4N$ & $2$ & 1185 & 22.74 & 24.96 \\
        $1 + 4N$ & $4$ & 2337 & 23.18 & 25.42 \\
        \rowcolor{gray!10}
        $\mathbf{1 + 4N}$ & $\mathbf{6}$ & \textbf{3489} & \textbf{23.52} & \textbf{25.87} \\
        $1 + 8N$ & $6$ & 3521 & 23.47 & 25.79 \\
        \bottomrule
    \end{tabular}
\end{table}